\documentclass[runningheads]{llncs}
\usepackage{graphicx}
\usepackage{latexsym}
\usepackage{microtype}
\usepackage{amssymb}
\usepackage{amsfonts}
\usepackage{amsmath} 
\usepackage{booktabs}
\usepackage{enumerate}
\usepackage{graphicx}
\usepackage{subfigure}
\usepackage{xspace}
\usepackage{float}
\usepackage{bbm}
\usepackage{bm}
\usepackage{multirow}
\usepackage{booktabs}
\usepackage{color}
\usepackage{framed}
\usepackage{stfloats}
\usepackage{iitem}
\usepackage{makecell}
\usepackage{float}
\restylefloat{table}
\usepackage{placeins}
\usepackage{afterpage}
\usepackage[normalem]{ulem}

\useunder{\uline}{\ul}{}
\usepackage{url}
\newcommand{\paratitle}[1]{\vspace{1.5ex}\noindent\textbf{#1}}
\newcommand{\ie}{\emph{i.e.,}\xspace}

\newcommand{\eg}{\emph{e.g.,}\xspace}

\newcommand{\ignore}[1]{}

\usepackage{pifont}
\newcommand{\cmark}{\ding{51}}%

\usepackage[noabbrev,capitalize]{cleveref} 

\crefname{equation}{equation}{equations}   
\crefname{footnote}{footnote}{footnotes}   
\crefname{line}{line}{lines}               
\crefname{section}{\S}{\S\S}
\Crefname{section}{\S}{\S\S}    
\crefformat{section}{#2\S#1#3}  
\Crefformat{section}{#2\S#1#3}
\crefrangeformat{section}{\S\S#3#1#4--#5#2#6}
\Crefrangeformat{section}{\S\S#3#1#4--#5#2#6}

\usepackage{xcolor}

\usepackage[misc]{ifsym}
\usepackage[marginal]{footmisc}

\begin{document}
\title{MASTER: Multi-task Pre-trained Bottlenecked Masked Autoencoders are Better Dense Retrievers}
\titlerunning{Multi-task Pre-trained Bottlenecked Masked Autoencoders}
%
\author{
Kun Zhou\inst{1,3}\orcidID{0000-0003-0650-9521}*
\and Xiao Liu\inst{4}\orcidID{0000-0002-8893-366X}
\and \\ Yeyun Gong\inst{4}\orcidID{0000-0001-9954-9674} 
\and Wayne Xin Zhao\inst{2,3}\orcidID{0000-0002-8333-6196}\Letter
\and \\ Daxin Jiang\inst{4}\orcidID{0000-0002-6657-5806} 
\and Nan Duan\inst{4}\orcidID{0000-0002-3387-4674} 
\and \\ Ji-Rong Wen\inst{1,2,3}\orcidID{0000-0002-9777-9676}
}
\institute{
School of Information, Renmin University of China 
\and Gaoling School of Artificial Intelligence, Renmin University of China
\and Beijing Key Laboratory of Big Data Management and Analysis Methods
\and Microsoft Research \\
\email{
francis\_kun\_zhou@163.com,
xiaoliu2@microsoft.com,
yegong@microsoft.com,
batmanfly@gmail.com,
nanduan@microsoft.com,
jrwen@ruc.edu.cn
}
}
%
\authorrunning{K. Zhou et al.}
%
%
\maketitle              
\begin{abstract}
\footnote{\textsuperscript{*}\:This work was done during internship at MSRA.}
\footnote{\textsuperscript{\Letter}\:Corresponding Author}
Pre-trained Transformers (\eg BERT) have been commonly used in existing dense retrieval methods for parameter initialization, and recent studies are exploring more effective pre-training tasks for further improving the quality of dense vectors.
Although various novel and effective tasks have been proposed, their different input formats and learning objectives make them hard to be integrated for jointly improving the model performance.
In this work, we aim to unify a variety of pre-training tasks into the bottlenecked masked autoencoder manner, and integrate them into a multi-task pre-trained model, namely MASTER.
Concretely, MASTER utilizes a shared-encoder multi-decoder architecture that can construct a representation bottleneck to compress the abundant semantic information across tasks into dense vectors.
Based on it, we integrate three types of representative pre-training tasks: corrupted passages recovering, related passages recovering and PLMs outputs recovering, to characterize the inner-passage information, inter-passage relations and PLMs knowledge.
Extensive experiments have shown that our approach outperforms competitive dense retrieval methods.
Our code and data are publicly released in \url{https://github.com/microsoft/SimXNS}.

\end{abstract}

\section{Introduction}
Recent years have witnessed the great success of dense retrieval methods~\cite{DBLP:conf/emnlp/KarpukhinOMLWEC20,zhou2022debiased,zhou2022simans,zhao2022dense} in industrial applications, \eg web search~\cite{zhou2022simans,Gao2021CondenserAP} and question answering~\cite{DBLP:conf/emnlp/KarpukhinOMLWEC20,Qu2021RocketQAAO}.
These methods typically encode queries and passages into low-dimensional dense vectors and utilize the vector similarity between them to measure semantic relevance.
In real-world applications, the dense vectors of a large number of passages will be pre-computed.
Then the approximate nearest neighbor (ANN) search techniques~\cite{DBLP:journals/tbd/JohnsonDJ21} can be incorporated for efficient retrieval.

In existing dense retrieval methods, pre-trained language models (PLMs)~\cite{devlin2019bert,zhao2023survey} have been widely adopted as the backbone, showing the superiority to generate high-quality dense vectors.
However, general PLMs (\eg BERT~\cite{devlin2019bert}) may not be the best for dense retrieval, as their produced native dense representations (usually the \texttt{[CLS]} embedding) are not designed on purpose to compress the information from the input text.
To solve it, recent studies~\cite{Gao2021CondenserAP,wu2023cot,ma2023cot} continually pre-train PLMs for improving the \texttt{[CLS]} embedding. 
Typically, they mainly focus on capturing the inner-passage information (\eg recovering masked tokens)~\cite{Liu2022RetroMAEPR,DBLP:journals/corr/abs-2207-02578} or inter-passage relations (\eg co-occurring passages)~\cite{DBLP:journals/corr/abs-2208-07670}, and specially design pre-training tasks.
After pre-training, the enhanced \texttt{[CLS]} embeddings would be fine-tuned on downstream passage retrieval tasks, achieving faster convergence and better performance than general PLMs.


\begin{figure}[t]
\centering
\includegraphics[width=\textwidth]{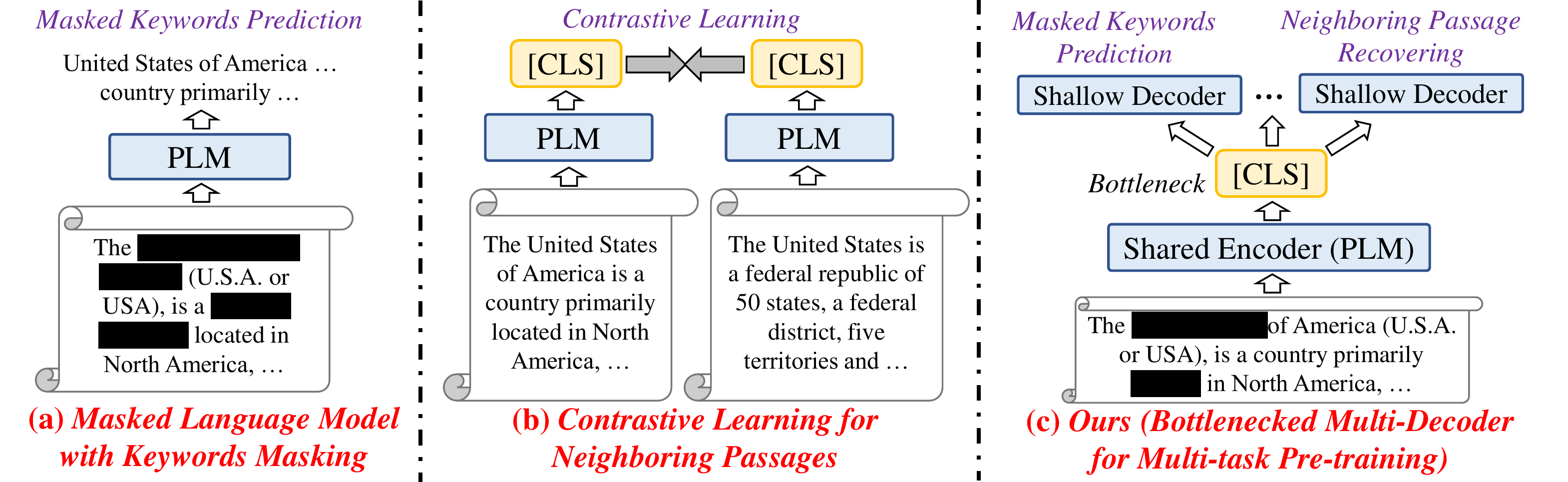}
\caption{The comparison of two representative pre-training tasks and our approach. Ours incorporates a bottlenecked multi-decoder architecture that unifies different tasks into the same input format and leverages specific decoders to deal with them separately.}
\label{fig:intro}
\end{figure}

As existing work has shown the effectiveness of capturing the two types of characteristics during pre-training by specific tasks, it is promising to combine these tasks for enhancing the \texttt{[CLS]} embedding.
Intuitively, by incorporating more tasks to capture more specific useful information, the \texttt{[CLS]} embedding would be further enriched during pre-training, helping it generalize better into downstream retrieval tasks.
However, due to the divergence of focused characteristics, the available pre-training tasks in existing work may adopt different settings in training objectives and input formats, \eg contrastive learning with co-occurring passages as positives and sampled negatives, and masked language model with special masking strategies on passages.
Such differences make it hard to combine existing pre-training tasks, and an arbitrary integration of these tasks may even cause detrimental interference in the semantics of the \texttt{[CLS]} embedding, leading to performance degradation.

To address this problem, we consider integrating multiple pre-training tasks in a unified input format and reducing the divergence of different training objectives.
Since most of the NLP tasks can be reformulated as the text-to-text format~\cite{raffel2020exploring}, we can also reconstruct the available pre-training tasks into such a format.
Concretely, the tasks for capturing the inner-passage information or inter-passage relations, can be converted as predicting the inner- or inter-passage textual information (\eg tokens) based on the same input passage.
Therefore, we can propose a unified framework for these tasks that adopts the PLM as the shared encoder, and multiple task-specific decoders.
As shown in Figure~\ref{fig:intro}, the shared encoder produces the \texttt{[CLS]} embedding for the input passage, and all the decoders mainly rely on the embedding for predicting the specific texts.
Such a way constructs an information-bottleneck architecture~\cite{Liu2022RetroMAEPR,DBLP:journals/corr/abs-2207-02578,DBLP:journals/corr/abs-2208-07670} where the PLM encoder is forced to inject sufficient task-specific information into the \texttt{[CLS]} embedding, for well accomplishing the tasks in decoders.

The proposed bottlenecked multi-decoder architecture provides a flexible way to integrating multiple different tasks for pre-training dense retriever.
Based on it, we can combine a diverse range of available tasks for capturing the useful information or relations from different perspectives.
Besides the commonly-used inner-passage information and inter-passage relations, we also consider to learn the knowledge from other public generative PLMs (\eg GPT-2~\cite{radford2019language}), for capturing useful information beyond the corpus.
Specifically, we devise three types of pre-training tasks for recovering corrupted passages, related passages, and PLMs output, respectively, including a total of five tasks in five decoders.
Inspired by the masked autoencoder method~(MAE)~\cite{he2022masked}, we perform aggressively masking on the decoders (\eg masking 50\% tokens), hence the deep encoder would be forced to generate compressed high-quality representations to recover them. 
Finally, we propose \textbf{MASTER}, a \textbf{m}ulti-t\textbf{as}k pre-\textbf{t}rained bottlenecked masked auto\textbf{e}ncode\textbf{r}, that adopts a shared-encoder multi-decoder architecture to integrate the five pre-training tasks in the bottlenecked MAE format.
To verify the effectiveness of our approach, we conduct extensive experiments on several text retrieval datasets.
Experimental results show that our approach can outperform competitive baselines.

\section{Related Work}

\paratitle{Dense Retrieval.}
Dense retrieval approaches \cite{DBLP:conf/emnlp/KarpukhinOMLWEC20} typically map queries and documents into low-dimensional dense vectors for evaluating their relevance, which support the efficient approximate nearest neighbor (ANN) search engines, \eg FAISS~\cite{DBLP:journals/tbd/JohnsonDJ21}.
For effectively training dense retrieval models, existing work typically leverages pre-trained Transformers~\cite{devlin2019bert} to initialize the dual encoders and then samples high-quality negatives for fine-tuning the encoders.
Early work~\cite{DBLP:conf/emnlp/KarpukhinOMLWEC20} mainly relies on in-batch random negatives and hard negatives mined by BM25.
Afterward, a line of work~\cite{Qu2021RocketQAAO,DBLP:conf/iclr/XiongXLTLBAO21} picks top-$k$ ranked documents by a trained dense retriever as hard negatives and improves the performance.
However, a common problem for such top-$k$ negative sampling strategies is that they are easy to select false negatives, which impedes better performances.
To alleviate it, current studies have explored several practical directions, \eg knowledge distillation~\cite{Qu2021RocketQAAO,sun2022lead,lin2023prod}, pre-training ~\cite{Gao2021CondenserAP} and negative sampling~\cite{zhou2022simans}.
Besides, recent work is also exploring more efficient and effective ways for training dense retrievers, \eg ambiguous negative sampling~\cite{zhou2022simans} and neural corpus index~\cite{zhou2023dynamicretriever}.

\paratitle{Pre-training for Dense Retrieval.}
As general PLMs~\cite{devlin2019bert} are pre-trained without any prior task knowledge, they are not ready to use for dense retrieval~\cite{Gao2021CondenserAP,DBLP:conf/acl/GaoC22}, especially in low-data situations.
To solve this issue, several studies~\cite{Gao2021CondenserAP,DBLP:journals/corr/abs-2208-07670} are proposed to make the output sentence embedding more informative and discriminative.
A type of work relies on the explicit relations between text pairs and designs the pre-training tasks based on the contrastive learning objective~\cite{DBLP:conf/acl/GaoC22,DBLP:conf/acl/RenLQLZSWWW21}, \eg inverse cloze task and contrastive span prediction.
Another line of work aim to compress the semantic information into the \texttt{[CLS]} embedding.
They leverage the masked autoencoder architecture that incorporates a deep encoder and a shallow decoder, forcing the \texttt{[CLS]} embedding of the input text from the encoders to recover itself~\cite{Liu2022RetroMAEPR,DBLP:journals/corr/abs-2208-07670,xiao2022retromae} or related texts~\cite{DBLP:journals/corr/abs-2207-02578}.

\section{Preliminary}


\paratitle{Task Definition.}
Given a query $q$, the dense retrieval task aims to retrieve the most relevant top-$k$ passages $\{p_{i}\}^{k}_{i=1}$ from a large candidate pool $\mathcal{P}$.
To achieve it, the dual-encoder architecture is widely used.
It consists of a query encoder $E_{q}$ and a passage encoder $E_{p}$, mapping the query $q$ and passage $p$ into $k$-dimensional dense vectors $\textbf{h}_{q}$ and $\textbf{h}_{p}$, respectively.
Then, the semantic relevance score of $q$ and $p$ will be computed using dot product as
\begin{equation}
    s(q,p)=\textbf{h}_{q} \cdot \textbf{h}_{p}.
\end{equation}

Existing work mostly adopts pre-trained Transformers (\eg BERT~\cite{devlin2019bert}) as the two encoders, using the representations of the \texttt{[CLS]} token as the dense vectors.
In this work, we aim to propose a more effective multi-task pre-training framework specially for the dense retrieval task, which learns to compress more useful information into the \texttt{[CLS]} representations.
Formally, given a pre-training corpus and a Transformer encoder, we focus on devising several tasks to pre-train the parameters of it.
Then, the pre-trained Transformer will be used as the backbone of the query encoder $E_{q}$ and passage encoder $E_{p}$, and can be fine-tuned on downstream dense retrieval tasks.

\paratitle{Fine-tuning Dense Retrievers.}
In the fine-tuning stage, the learning objective is to pull the representations of a query $q$ and its relevant passages $\mathcal{P}^{+}$ together (as positives), while pushing apart irrelevant ones $\mathcal{P}^{-}=\mathcal{P}\setminus \mathcal{P}^{+}$ (as negatives).
Therefore, high-quality negatives are critical to the effectiveness of dense retrievers.
Existing work commonly leverages the BM25 negatives \cite{DBLP:conf/emnlp/KarpukhinOMLWEC20} or the top-$k$ ranked negatives mined by a well-trained dense retriever \cite{Qu2021RocketQAAO,DBLP:conf/iclr/XiongXLTLBAO21}, denoted as $\tilde{\mathcal{D}}^{-}$.
Then, the optimization objective can be formulated as:
\begin{equation}
\small
    \theta^{*}=\arg\min_{\theta} \sum_{q}\sum_{d^{+}\in\mathcal{D}^{+}}\sum_{d^{-}\in\tilde{\mathcal{D}}^{-}} l(s(q,d^{+}),s(q,d^{-})), 
\end{equation}
where $l(\cdot)$ is the loss function.
Besides, as the top-$k$ hard negatives may contain false negatives, recent studies~\cite{Qu2021RocketQAAO,DBLP:conf/emnlp/RenQLZSWWW21,DBLP:journals/corr/abs-2205-09153} have adopted knowledge distillation strategies to solve it.
They rely on pre-learned cross-encoder rerankers to produce soft labels on $\tilde{\mathcal{D}}^{-}$, and minimize the KL divergence between the dual encoders' outputs and the soft labels.

\section{Approach}

\begin{figure}[t]
\centering
\includegraphics[width=0.95\textwidth]{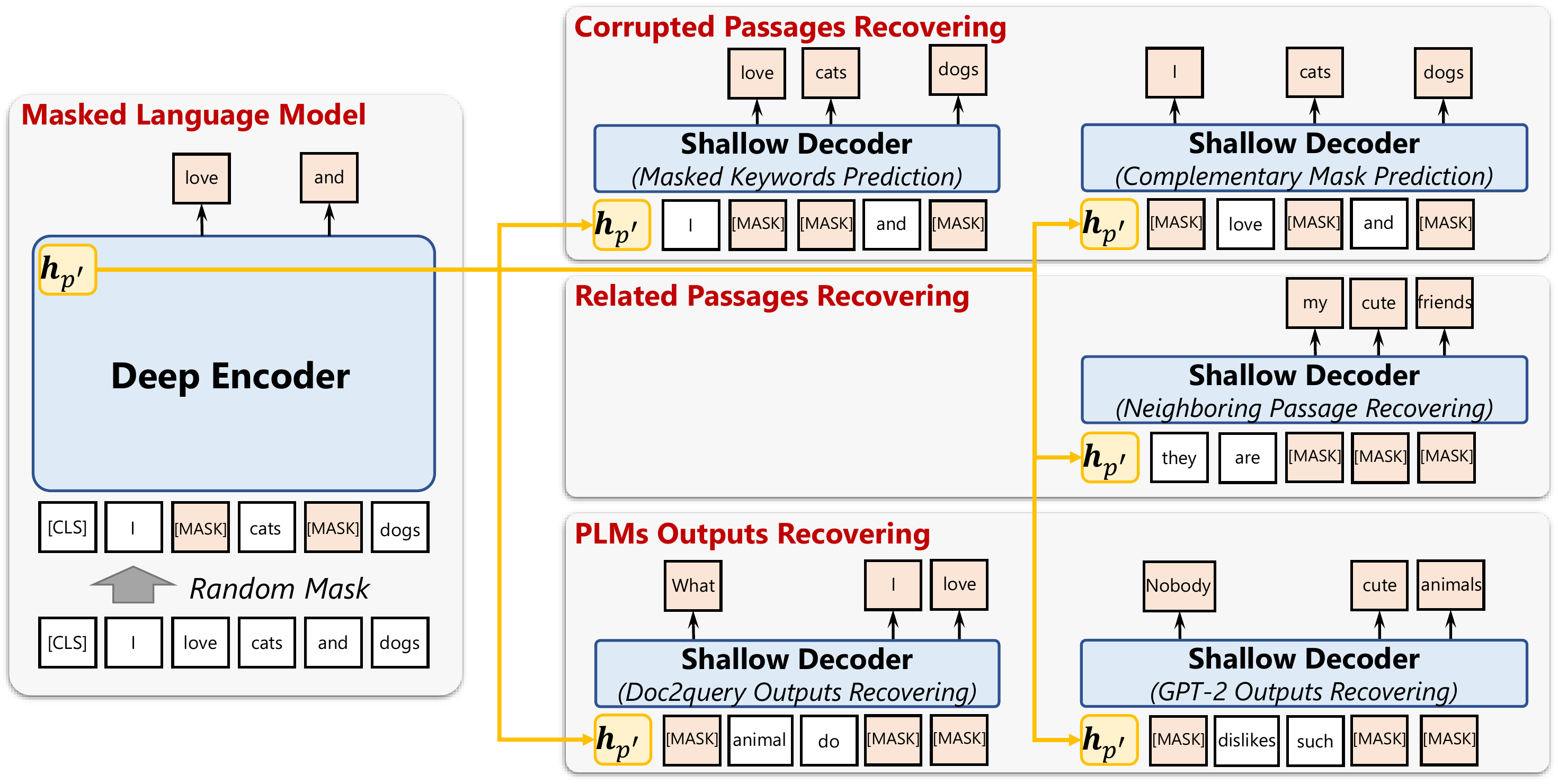}
\caption{The overview of MASTER. We adopt a bottlenecked multi-decoder architecture, and design three types of pre-training tasks, totally five decoders for specific tasks.}
\label{fig:framework}
\end{figure}

In this section, we present MASTER, an approach to pre-training an effective dense retriever.
We first introduce the bottlenecked model architecture (consisting of a PLM encoder and multiple shallow decoders), then describe our adopted three types of pre-training tasks unified as the bottlenecked masked autoencoding manner.
\cref{fig:framework} shows the overview of our approach.

\subsection{Bottlenecked Multi-Decoder Architecture}
To pre-train the dense retriever for compressing useful information into the dense vectors, we design a bottlenecked multi-decoder architecture.
In the architecture, we incorporate a deep Transformer encoder to compress the input text into a dense vector, and five shallow decoders corresponding to different pre-training tasks to capture diverse semantics and relations.

Concretely, the deep Transformer encoder shares the same architecture as BERT~\cite{devlin2019bert}, and can be initialized with its pre-trained parameters.
Given a passage $p$ from the pre-training corpus, we leverage the deep encoder to encode it, and select the output representation of the \texttt{[CLS]} token as its dense vector $\mathbf{h}_p$.
Following existing work~\cite{Gao2021CondenserAP,Lu2021LessIM}, we employ a masked language model task to pre-train the encoder. 
Formally, a certain percentage $\alpha\%$ of tokens from $p$ will be masked to obtain $p'$, and the encoder needs to predict them as:
\begin{equation}
L_{\textrm{MLM}}=\sum_{t_{i}\in \mathcal{M}_{p'}} -\log p(t_{i}|p';\Theta_{E})
\end{equation}
where $\mathcal{M}_{p'}$ denotes the masked tokens in $p'$, $\Theta_{E}$ denotes the parameters of the encoder.
The multiple shallow decoders are all the 2-layer bi-directional Transformer, and share the embedding matrix and language modeling head with the deep encoder.
For each decoder, its input is an aggressive masked text $x'$ (masking rate $\beta \geq 50\%$) that requires to be recovered.
Besides, the dense vector $\mathbf{h}_{p'}$ from the encoder will be inserted into the decoder to replace the original \texttt{[CLS]} token embedding.
In this way, the learning objective of each decoder is:
\begin{equation}
    L_{D}=\sum_{t_{i}\in \mathcal{M}_{x'}} -\log p(t_{i}|x',\mathbf{h}_{p'};\Theta_{E},\Theta_{D})
\end{equation}
where $\mathcal{M}_{x'}$ denotes the masked tokens in $x'$, $\Theta_{D}$ denotes the parameters of the decoder.
Such a way builds the information bottleneck where multiple decoders rely on $\mathbf{h}_{p'}$ to recover the input, forcing it to reserve more useful information.

\subsection{Multi-Task Pre-training}
Based on the architecture, we devise multiple pre-training tasks, to help dense vectors capture more useful information.
Concretely, we adopt three types of tasks to capture the semantic information within passages, relations with other passages, and knowledge from other PLMs, namely corrupted passages recovering, related passages recovering and PLMs outputs recovering, respectively.

\paratitle{Corrupted Passages Recovering.}
Given a passage $p$ from the pre-training corpus, the corrupted passages recovering tasks~(CPR) first mask its contained tokens to compose the inputs of the encoder $p'$ and decoder $\hat{p}'$ according to the mask rates $\alpha\%$ and $\beta\%$ respectively.
Then, the output dense vector $\mathbf{h}_{p'}$ from the encoder will be leveraged to help the shallow decoder to recover $\hat{p}'$ into $p$.
Such a way is helpful to compress important semantic information from the passage into the dense vector.
To achieve it, we design two pre-training tasks by utilizing special masking mechanisms for the decoder, namely masked keywords prediction (MKP) and complementary mask prediction (CMP).

For MKP, we aim to mask more keywords in the decoder, as they may reflect important semantic information of the passage.
Specifically, we rely on the TF-IDF weights~\cite{ramos2003using} to obtain a masked probability distribution about words in the passage, where keywords with low frequencies would receive larger probabilities to be masked. 
In this way, the input masked passage $\hat{p}'_{\textrm{MKP}}$ of the decoder will lose most keywords, which will force the dense vector $\mathbf{h}_{p'}$ to well reserve their information for recovering.
For CMP, given the passage $p$, we leverage a complementary mask mechanism in the decoder that masks the unmasked tokens from the input of the encoder $p'$.
As a result, the incomplete inputs of the encoder and decoder will be complementary, and the dense vector $\mathbf{h}_{p'}$ should accurately remember all the unmasked input information from $p'$ for recovering $\hat{p}'_{\textrm{CMP}}$.

Finally, the pre-training objective of the CPR tasks is given by combining the above two tasks as:
\begin{align}
\small
    L_{\textrm{CPR}}&=\sum_{t_{i}\in \mathcal{M}_{\textrm{MKP}}} -\log p(t_{i}|\hat{p}'_{\textrm{MKP}},\mathbf{h}_{p'};\Theta_{E},\Theta^{\textrm{MKP}}_{D}) \nonumber \\
    &+\sum_{t_{i}\in \mathcal{M}_{\textrm{CMP}}} -\log p(t_{i}|\hat{p}'_{\textrm{CMP}},\mathbf{h}_{p'};\Theta_{E},\Theta^{\textrm{CMP}}_{D}), \nonumber
\end{align}
where $\mathcal{M}_{\textrm{MKP}}$ and $\mathcal{M}_{\textrm{CMP}}$ denote the masked tokens in $\hat{p}'_{\textrm{MKP}}$ and $\hat{p}'_{\textrm{CMP}}$, respectively, and $\Theta^{\textrm{MKP}}_{D}$ and $\Theta^{\textrm{CMP}}_{D}$ are the parameters of the two specific decoders.

\paratitle{Related Passages Recovering.}
The related passages recovering task~(RPR) aims to model the semantic relationships between related passages.
In this work, we focus on the commonly-used and easily-obtained co-occurrence relation from the pre-training corpus.
Based on this motivation, we collect the passage pairs $\{\langle p_{i}, p_{i+1} \rangle\}$ that are neighbouring spans in a document, and devise the neighbouring passage recovering task (NPR).

In NPR, given a neighbouring passage pair $\langle p_{i}, p_{i+1} \rangle$, we rely on the mask rates $\alpha\%$ and $\beta\%$ to mask their tokens for composing the inputs of the encoder $p'_{i}$ and decoder $p'_{i+1}$, respectively.
Next, the output dense vector of $p'_{i}$ from the deep encoder is utilized to help the decoder recover $p'_{i+1}$.
Such a way encourages the dense vector to retain the information related to the neighbouring passage, capturing the intrinsic token-level correlations across the two passages.
Besides, we also rely on the TF-IDF weights of words to mask more keywords in the decoder as MKP, which further increases the difficulty of this task and forces the dense vector to focus more on the key information.
The learning objective of the RPR task can be defined as:
\begin{equation}
    L_{\textrm{RPR}}=\sum_{t_{i}\in \mathcal{M}_{\textrm{NPR}}} -\log p(t_{i}|p'_{i+1},\mathbf{h}_{p'_{i}};\Theta_{E},\Theta^{\textrm{NPR}}_{D}), \nonumber
\end{equation}
where $\mathcal{M}_{\textrm{NPR}}$ and $\Theta^{\textrm{NPR}}_{D}$ denote the masked tokens in $p'_{i+1}$ and the parameters of the decoder specially for the NPR task, respectively.
Note that existing work~\cite{lee2019latent,Ma2022PretrainAD} has also considered the neighbouring relations and mostly adopts the contrastive learning objective to capture it.
In fact, contrastive learning mainly aims to characterize the passage-level semantics and would be affected by the quality of sampled negative passages.
As a comparison, the NPR task can capture more fine-grained token-level characteristics, and such a generative way only focuses on modeling the relations between neighbouring passages, avoiding the influence from other passages.

\paratitle{PLMs Outputs Recovering.}
The above tasks are able to capture the semantic information and relations within the unsupervised pre-training corpus.
We further consider to learn the knowledge from other PLMs, to capture more rich information beyond the corpus.
Based on this idea, we design the PLMs outputs recovering tasks~(POR) that aim to recover the outputs of two generative PLMs, consisting of the doc2query outputs recovering~(DOR)  and GPT-2 outputs recovering~(GOR) tasks.

Given a passage $p$, we leverage a public well-trained doc2query model~\cite{nogueira2019doc2query} to generate $k$ relevant queries $\{q_{i}\}_{i=1}^{k}$ and concatenate them into a long sentence $s_{(q)}$, as the generated queries have shown effectiveness in previous dense retrieval methods~\cite{DBLP:journals/corr/abs-1904-08375}.
Besides, we also use $p$ as the prompt to guide the popular autoregressive GPT-2 model~\cite{radford2019language} to generate a long sentence $s_{(g)}$, as GPT-2 has shown surprising performance in generating informative long text.
Then, we aggressively mask the tokens in $s_{(q)}$ and $s_{(g)}$ according to the mask rate $\beta\%$, to obtain the inputs $s'_{(q)}$ and $s'_{(g)}$ of two task-specific decoders.
Similar to above tasks, the two decoders also rely on the dense vector $\mathbf{h}_{p'}$ to recover the generated texts, and the pre-training objective of the POR tasks is the combination of the two tasks as:
\begin{align}
\small
    L_{\textrm{POR}}&=\sum_{t_{i}\in \mathcal{M}_{\textrm{DOR}}} -\log p(t_{i}|s'_{(q)},\mathbf{h}_{p'};\Theta_{E},\Theta^{\textrm{DOR}}_{D}) \nonumber \\
    &+\sum_{t_{i}\in \mathcal{M}_{\textrm{GOR}}} -\log p(t_{i}|s'_{(g)},\mathbf{h}_{p'};\Theta_{E},\Theta^{\textrm{GOR}}_{D}), \nonumber
\end{align}
where $\mathcal{M}_{\textrm{DOR}}$ and $\mathcal{M}_{\textrm{GOR}}$ denote the masked tokens in $s'_{(q)}$ and $s'_{(g)}$, respectively, and $\Theta^{\textrm{DOR}}_{D}$ and $\Theta^{\textrm{GOR}}_{D}$ are the parameters of the two specific decoders, respectively.
In this way, the dense vector is enhanced to capture richer knowledge from other PLMs, and learn more information not included in the corpus.
Such a way is similar to the knowledge distillation process that transfers the learned knowledge from PLMs into the dense vector by forcing it to predict the PLMs' outputs.

\subsection{Learning}
During pre-training, we optimize the parameters in the deep encoder and the multiple shallow decoders using the above pre-training tasks, denoted as:
\begin{equation}
    L_{\textrm{total}}=L_{\textrm{MLM}}+L_{\textrm{CPR}}+L_{\textrm{RPR}}+L_{\textrm{POR}}
\end{equation}
During fine-tuning, we utilize the pre-trained deep encoder as the backbone of the query and passage encoders.
Following the pipeline in previous dense retrieval methods~\cite{DBLP:conf/acl/GaoC22,DBLP:journals/corr/abs-2207-02578,DBLP:journals/corr/abs-2208-07670}, we first train the \textbf{Retriever$_{1}$}  using the in-batch negatives and BM25 hard negatives.
Then, we utilize Retriever$_{1}$ to mine hard negatives from a large-scale passage pool, and leverage these negatives and in-batch negatives to train the \textbf{Retriever$_{2}$}.
Next, we train a cross-encoder reranker model based on the mined negatives from Retriever$_{2}$.
Finally, we distil the knowledge from the reranker into the \textbf{Retriever$_{\textrm{distil}}$} by using it to produce soft labels for both positives and mined negatives from Retriever$_{2}$.
Note that our pre-trained encoder is used to initialize the Retriever$_{1}$, Retriever$_{2}$ and Retriever$_{\textrm{distil}}$.


\begin{table}[t]
\caption{Statistics of the text retrieval datasets.}
\setlength\tabcolsep{3pt}
\centering
\small
\begin{tabular}{l|cccc}
\hline
\textbf{Dataset} & \textbf{Train} & \textbf{Dev} & \textbf{Test} & \textbf{\#Passage} \\
\hline
MS MARCO Passage Ranking~(MS-Pas) & 502,939 & 6,980 & - & 8,841,823 \\
TREC 2019 Deep Learning Track~(TREC-2019) & - & - & 200 & 8,841,823 \\
TREC-2020 Deep Learning Track~(TREC-2020) & - & - & 200 & 8,841,823 \\
Natural Questions~(NQ) & 58,880 & 8,757 & 3,610 & 21,015,324 \\
\hline
\end{tabular}
\label{tab:dataset}
\end{table}

\section{Experiment}

\subsection{Experimental Setting}

\paratitle{Datasets and Evaluation.}
We conduct experiments on several text retrieval datasets: 
MS-MARCO \cite{DBLP:conf/nips/NguyenRSGTMD16},
TREC-2019 Deep Learning Track \cite{DBLP:journals/corr/abs-2003-07820},
TREC-2020 Deep Learning Track \cite{DBLP:journals/corr/abs-2102-07662},
and Natural Questions (\textsc{NQ}) \cite{DBLP:journals/tacl/KwiatkowskiPRCP19}.
The statistics of the above datasets are shown in \cref{tab:dataset}.
MS-MARCO consists of real queries collected from Bing search engine.
NQ is an open domain QA dataset.

\paratitle{Baselines.}
We compare our approach with a variety of methods:
BM25~\cite{yang2017anserini} is a widely-used sparse retriever based on exact matching. DeepCT~\cite{dai2019deeper} and docT5query~\cite{nogueira2019doc2query} enhance BM25 with neural models.
ANCE~\cite{DBLP:conf/iclr/XiongXLTLBAO21}, TAS-B~\cite{DBLP:conf/sigir/HofstatterLYLH21} and STAR~\cite{DBLP:conf/sigir/ZhanM0G0M21} are dense retrieval methods that adopt top-$k$ hard negatives to improve training.
RocketQA~\cite{Qu2021RocketQAAO}, AR2~\cite{DBLP:conf/iclr/ZhangGS0DC22} and ERNIE-search~\cite{DBLP:journals/corr/abs-2205-09153} utilize knowledge distillation technique that leverages a teacher model to guide the training of the dual-encoder retriever.
COIL~\cite{Gao2021COILRE}, ColBERT~\cite{DBLP:conf/sigir/KhattabZ20} and ColBERTv2~\cite{DBLP:conf/naacl/SanthanamKSPZ22} utilize multiple representations for text retrieval.
SEED~\cite{Lu2021LessIM}, RetroMAE~\cite{Liu2022RetroMAEPR}, Condenser~\cite{gao2021your},
PAIR~\cite{DBLP:conf/acl/RenLQLZSWWW21}, coCondenser~\cite{DBLP:conf/acl/GaoC22}, CoT-MAE~\cite{DBLP:journals/corr/abs-2208-07670} and SimLM~\cite{DBLP:journals/corr/abs-2207-02578} design special pre-training tasks to improve the backbone models.

\paratitle{Implementation Details.}
During pre-training, we leverage BERT-base to initialize the shared encoder, and all decoders are randomly initialized two-layer Transformers.
Following previous work~\cite{DBLP:conf/acl/GaoC22,DBLP:journals/corr/abs-2208-07670,DBLP:journals/corr/abs-2207-02578}, we leverage the passages in MS-MARCO and NQ dataset as the pre-training corpus of them, respectively. 
The pre-training steps are setting to 120k.
During fine-tuning, we also follow SimLM that progressively trains Retriever$_{1}$, Retriever$_{2}$, and Retriever$_{\textrm{distil}}$, where our pre-trained deep Transformer encoder is leveraged to initialize their parameters.
Our all other hyper-parameters are the same as SimLM~\cite{DBLP:journals/corr/abs-2207-02578}. 

\begin{table}[t]
\small
\centering
\caption{Results on three web search datasets.
The best and second-best methods are marked in bold and underlined, respectively.
The \cmark in the column of ``with KD?'' means that the model has used knowledge distillation.
}
\resizebox{\textwidth}{!}{
\begin{tabular}{@{}l|c|ccc|c|c@{}}
\hline
\multirow{2}{*}{\textbf{Model}} & \multirow{2}{*}{\textbf{with KD?}} & \multicolumn{3}{c|}{\textbf{\textsc{MS-MARCO}}} & \textbf{\textsc{TREC-19}} & \textbf{\textsc{TREC-20}} \\
 & & \textbf{MRR@10} & \textbf{R@50} & \textbf{R@1k} & \textbf{nDCG@10} & \textbf{nDCG@10} \\ \hline
BM25~\cite{yang2017anserini} &  & 18.5 & 58.5 & 85.7 & 51.2 & 47.7 \\
DeepCT~\cite{dai2019deeper} &  & 24.3 & 69.0 & 91.0 & 57.2 & - \\
docT5query~\cite{nogueira2019doc2query} &  & 27.7 & 75.6 & 94.7 & 64.2 & - \\ \hline
ANCE~\cite{DBLP:conf/iclr/XiongXLTLBAO21} &  & 33.0 & - & 95.9 & 64.5 & 64.6 \\
STAR~\cite{DBLP:conf/sigir/ZhanM0G0M21} &  & 34.7 & - & - & 68.3 & - \\
TAS-B~\cite{DBLP:conf/sigir/HofstatterLYLH21} & \cmark  & 34.0 & - & 97.5 & 71.2 & 69.3 \\
RocketQA~\cite{Qu2021RocketQAAO} & \cmark  & 37.0 & 85.5 & 97.9 & - & - \\
RocketQAv2~\cite{DBLP:conf/emnlp/RenQLZSWWW21} & \cmark  & 38.8 & 86.2 & 98.1 & - & - \\
AR2~\cite{DBLP:conf/iclr/ZhangGS0DC22} & \cmark  & 39.5 & 87.8 & 98.6 & - & - \\
ERNIE-Search~\cite{DBLP:journals/corr/abs-2205-09153} & \cmark  & 40.1 & 87.7 & 98.2 & - & - \\
AR2+SimANS~\cite{zhou2022simans} & \cmark & 40.9 & \textbf{88.7} &\underline{98.7} & - & - \\ \hline
COIL~\cite{Gao2021COILRE} & & 35.5 & - & 96.3 & 70.4 & - \\
ColBERT~\cite{DBLP:conf/sigir/KhattabZ20} & & 36.0 & 82.9 & 96.8 & - & - \\
ColBERTv2~\cite{DBLP:conf/naacl/SanthanamKSPZ22} & \cmark & 39.7 & 86.8 & 98.4 & - & - \\ \hline
SEED~\cite{Lu2021LessIM} & & 33.9 & - & 96.1 & - & - \\
RetroMAE~\cite{Liu2022RetroMAEPR} & & 35.0 & - & 97.6 & - & - \\
Condenser~\cite{Gao2021CondenserAP} & & 36.6 & - & 97.4 & 69.8 & - \\
coCondenser~\cite{DBLP:conf/acl/GaoC22} & & 38.2 & 86.5 & 98.4 & \underline{71.7} & 68.4 \\
CoT-MAE~\cite{DBLP:journals/corr/abs-2208-07670} & & 39.4 & 87.0 & \underline{98.7} & - & \underline{70.4} \\
PAIR~\cite{DBLP:conf/acl/RenLQLZSWWW21} & \cmark & 37.9 & 86.4 & 98.2 & - & - \\
SimLM~\cite{DBLP:journals/corr/abs-2207-02578} & \cmark & \underline{41.1} & 87.8 & \underline{98.7} & 71.2 & 69.7 \\ \hline
MASTER & \cmark & \textbf{41.2} & \underline{88.6} & \textbf{98.8} & \textbf{72.7} & \textbf{71.7}  \\ \hline
\end{tabular}
}
\label{tab:ir_results}
\end{table}

\subsection{Main Results}

\paratitle{Performance on Web Search Datasets.}
\cref{tab:ir_results} shows the results on three web search benchmarks, \ie MS-MARCO, TREC-2019 and TREC-2020.
First, we can see that with or without distillation strategy, the best baselines are both pre-training dense retrieval methods, \ie CoT-MAE and SimLM, even outperforming methods using multiple representations.
It indicates that proper pre-training strategies are helpful to the downstream dense passage retrieval tasks.
Second, SimLM mostly outperforms other baselines. It employs a bottlenecked architecture that learns to compress the input information into a dense vector, and adopts a replaced language modeling objective to pre-train it. Such a way is more effective to force the dense vector to reserve the important semantics.

Besides, our approach outperforms all the baselines in terms of all metrics on all datasets.
Our approach adopts a multi-task pre-training framework that unifies five tasks on recovering of corrupted passages, related passages and PLMs outputs, based on a bottlenecked one-encoder multi-decoder architecture. 
In this way, we can force the output dense vector from the encoder to be more informative and functional to accomplish these tasks, leading to better representative capacity.

\ignore{
\begin{table}[t]
\small
\centering
\setlength\tabcolsep{3pt}
\caption{Results on NQ. We report the performance of Retriever$_\text{2}$ without knowledge distillation.}
\begin{tabular}{@{}l|cc@{}}
\hline
\multirow{2}{*}{\textbf{Model}} & \multicolumn{2}{c}{\textbf{\textsc{NQ}}} \\
 & \textbf{R@20} & \textbf{R@100} \\ \hline
BM25~\cite{yang2017anserini} & 59.1 & 73.7 \\
DPR$_\text{single}$~\cite{DBLP:conf/emnlp/KarpukhinOMLWEC20} & 78.4 & 85.4 \\
ANCE~\cite{DBLP:conf/iclr/XiongXLTLBAO21} & 81.9 & 87.5 \\
RocketQA~\cite{Qu2021RocketQAAO} & 82.7 & 88.5 \\
RocketQAv2~\cite{DBLP:conf/emnlp/RenQLZSWWW21} & 83.7 & 89.0 \\ \hline
Condenser~\cite{Gao2021CondenserAP} & 83.2 & 88.4 \\
PAIR~\cite{DBLP:conf/acl/RenLQLZSWWW21} & 83.5 & 89.1 \\
coCondenser~\cite{DBLP:conf/acl/GaoC22} & 84.3 & 89.0 \\
SimLM~\cite{DBLP:journals/corr/abs-2207-02578} & 84.3 & 89.3 \\ \hline
MASTER & \textbf{84.6} & \textbf{89.4} \\ \hline
\end{tabular}
\label{tab:dpr_result}
\end{table}}

\begin{table}[t]
\small
\centering
\setlength\tabcolsep{3pt}
\caption{The performance of Retriever$_\text{2}$ without knowledge distillation on NQ.}
\begin{tabular}{@{}l|ccccccc|c@{}}
\hline
Model &  DPR & ANCE & RocketQA & Condenser & PAIR & coCondenser & SimLM & MASTER \\ \hline
R@20 & 78.4 & 81.9 & 82.7 & 83.2 & 83.5 & 84.3 & 84.3 & \textbf{84.6} \\ 
R@100 & 85.4 & 87.5 & 88.5 & 88.4 & 89.1 & 89.0 & 89.3 & \textbf{89.4} \\ \hline
\end{tabular}
\label{tab:dpr_result}
\end{table}

\paratitle{Performance on Open Domain QA Datasets.}
\cref{tab:dpr_result} shows the results an open domain QA datasets, NQ.
For a fair comparison, we only report the performance of Retriever$_2$ without performing knowledge distillation.
First, we can also see that pre-training dense retrieval methods mostly outperform other methods. It further indicates the effectiveness of pre-training techniques in open domain QA tasks.
Besides, coCondenser and SimLM perform better than other methods, the reason is that they both adopt a bottlenecked architecture to compress the information into the dense vectors.
Finally, we can see that our approach outperforms all the baselines.
As a comparison, our approach can enhance the informativeness of dense vectors by integrating multiple pre-training tasks, which compress the semantic information within passages, model the relations between passages, and learn the knowledge from other PLMs.

\begin{table}[t]
\centering
\small
\caption{Zero-shot dense retrieval nDCG@10 performances on \textsc{beir} benchmark. Results with * are from our reproduction.}
\resizebox{\textwidth}{!}{
\begin{tabular}{l|ccccccc|c}
\hline
\textbf{Dataset} & 
\textbf{BERT} & \textbf{LaPraDoR} & \textbf{SimCSE} & \textbf{DiffCSE} & \textbf{SEED} & \textbf{Condenser} & \textbf{SimLM*} & \textbf{MASTER} \\
\hline
TREC-COVID & 0.615  & 0.492 & 0.460 & 0.492 & 0.627 & \textbf{0.750} & 0.637 & 0.620 \\
BioASQ & 0.253  & 0.308 & 0.263 & 0.258 & 0.308 & 0.322 & 0.350 & \textbf{0.354} \\
NFCorpus & 0.260  & \textbf{0.335} & 0.260 & 0.259 & 0.278 & 0.277 & 0.323 & 0.330 \\
\hline
NQ & 0.467  & 0.473 & 0.435 & 0.412 & 0.446 & 0.486 & 0.477 & \textbf{0.516} \\
HotpotQA & 0.488  & 0.495 & 0.502 & 0.499 & 0.541 & 0.538 & 0.581 & \textbf{0.589} \\
FiQA-2018 & 0.252  & 0.314 & 0.250 & 0.229 & 0.259 & 0.259 & 0.292 & \textbf{0.328} \\
\hline
Signal-1M(RT) & 0.204  &  0.231 & \textbf{0.262} & 0.260 & 0.256 & 0.261 & 0.257 & 0.252 \\
\hline
TREC-NEWS & 0.362  & 0.374 & 0.356 & 0.363 & 0.358 & 0.376 & 0.326 & \textbf{0.409} \\
Robust04 & 0.351  & 0.368 & 0.330 & 0.343 & 0.365 & 0.349 & 0.368 & \textbf{0.405} \\
\hline
ArguAna & 0.265  & \textbf{0.469} & 0.413 & 0.468 & 0.389 & 0.298 & 0.421 & 0.395 \\
Touche-2020 & 0.259  &  0.182 & 0.159 & 0.168 & 0.225 & 0.248 & 0.292 & \textbf{0.320} \\
\hline
CQADupStack & 0.282  &  0.288 & 0.290 & 0.305 & 0.290 & \textbf{0.347} & 0.332 & 0.327 \\
Quora & 0.787  & 0.847 & 0.844 & 0.850 & 0.852 & \textbf{0.853} & 0.773 & 0.791 \\
\hline
DBPedia & 0.314  & 0.338 & 0.314 & 0.303 & 0.330 & 0.339 & 0.345 & \textbf{0.399} \\
\hline
SCIDOCS & 0.113  & \textbf{0.155} & 0.124 & 0.125 & 0.124 & 0.133 & 0.145 & 0.141 \\
\hline
FEVER & 0.682  & 0.646 & 0.623 & 0.641 & 0.641 & 0.691 & 0.657 & \textbf{0.692} \\
Climate-FEVER & 0.187  & 0.209 & 0.211 & 0.200 & 0.176 & 0.211 & 0.163 & \textbf{0.215} \\
SciFact & 0.533  & 0.599 & 0.554 & 0.523 & 0.575 & 0.593 & 0.588 & \textbf{0.637} \\
\hline
\textbf{Avg.} & 0.371  & 0.396 & 0.369 & 0.372 & 0.391 & 0.407 & 0.407 & \textbf{0.429} \\
\hline
\end{tabular}
}
\label{tab:beir_ndcg}
\end{table}

\paratitle{Zero-Shot Evaluation.} 
We evaluate the zero-shot retrieval performance of our approach on BEIR benchmark~\cite{DBLP:journals/corr/abs-2104-08663}. 
It contains 18 datasets, covering dense retrieval tasks across different domains.
Following~\cite{DBLP:journals/corr/abs-2104-08663}, we fine-tune our approach in MS-MARCO training set and evaluate it on the BEIR benchmark using the official evaluation toolkit.
nDCG@10 is chosen as the evaluation metrics.
As shown in \cref{tab:beir_ndcg}, the average performance of our approach surpasses all baselines significantly. 
Since our approach incorporates multiple pre-training tasks for learning the dense representations, such a way can enrich the informativeness of them and help better adapt into different domains and retrieval tasks.

\subsection{Further Analysis}

\paratitle{Fine-tuning Performance in Three Stages.}
To further investigate the effectiveness of our approach, we show the performances of MASTER and other pre-training dense retrieval methods in each stage of our fine-tuning pipeline.
Here, the models in the three stages are all initialized by corresponding pre-trained parameters of these methods.
As shown in \cref{tab:cocon_comparison}, the performances of all pre-training methods are consistently improving with the process of the three-stage training.
In addition, our approach also outperforms all other pre-training methods in the three stages.
It indicates the superiority of our proposed multi-task pre-training strategy.

\begin{table}[t]
\small
\centering
\caption{Comparison with different pre-training dense retrieval methods in three stages of our fine-tuning pipeline on the dev set of \textsc{MS-MARCO}.}
\resizebox{\textwidth}{!}{
\begin{tabular}{@{}l|cc|cc|cc|cc@{}}
\hline
\multirow{2}{*}{\textbf{Model}} & 
\multicolumn{2}{c|}{\textbf{coCondenser}} &
\multicolumn{2}{c|}{\textbf{CoTMAE}}   &
\multicolumn{2}{c|}{\textbf{SimLM}} &
\multicolumn{2}{c}{\textbf{MASTER}} \\
 & \textbf{MRR@10} & \textbf{R@1k} & \textbf{MRR@10} & \textbf{R@1k} & \textbf{MRR@10} & \textbf{R@1k} & \textbf{MRR@10} & \textbf{R@1k}\\ \hline
Retriever$_{1}$ & 35.7 & 97.8 & 36.8 & 98.3 & 38.0 & 98.3& \textbf{38.3} & \textbf{98.8} \\
Retriever$_{2}$ & 38.2 & 98.4 & 39.2 & 98.7 & 39.1 & 98.6 & \textbf{40.4} & \textbf{98.8} \\
Retriever$_{\textrm{distil}}$ & 40.2 & 98.3 & 40.4 & 98.7 & 41.1 & 98.7 & \textbf{41.2} & \textbf{98.8} \\ \hline
\end{tabular}
}
\label{tab:cocon_comparison}
\end{table}

\begin{table}[t]
\small
\centering
\caption{Ablation and variation study of our approach. We report MRR@10 of the retriever$_1$ and retriever$_2$ on the dev set of \textsc{MS-MARCO}.}
\begin{tabular}{l|c|ccccc}
\hline
\textbf{Model} & \textbf{MASTER} & \textbf{w/o CPR} & \textbf{w/o RPR} & \textbf{w/o POR} & \textbf{+Shared-Dec} & \textbf{SimLM} \\ \hline
Retriever$_{1}$ & \textbf{38.3} & 37.7 & 37.6 & 37.6 & 37.4 & 38.0 \\
Retriever$_{2}$ & \textbf{40.4} & 39.9 & 39.8 & 39.8 & 39.1 & 39.1 \\ \hline
\end{tabular}
\label{tab:objectives}
\end{table}

\paratitle{Ablation and Variation Study.}
Our proposed approach incorporates a multi-decoder architecture and three types of tasks for pre-training.
To verify the effectiveness of each part, we conduct the ablation and variation study on the dev set of MS-MARCO to analyze their contributions.
We remove the CPR, RPR and POR tasks individually, and propose a variants that adopts a shared decoder to deal with the multiple tasks.
As shown in \cref{tab:objectives}, we can see that all the ablation and variation models will lead to the performance degradation.
It indicates that all the pre-training tasks and our multi-decoder architecture are useful to improve the performance.
Besides, after removing any type of pre-training tasks, our Retriever$_{2}$ still outperforms the SOTA method, SimLM.
It further shows the promising effectiveness of multi-task pre-training for dense retrieval tasks.

\begin{figure}[t]
    \small
	\centering
	\subfigure{\includegraphics[width=0.48\textwidth]{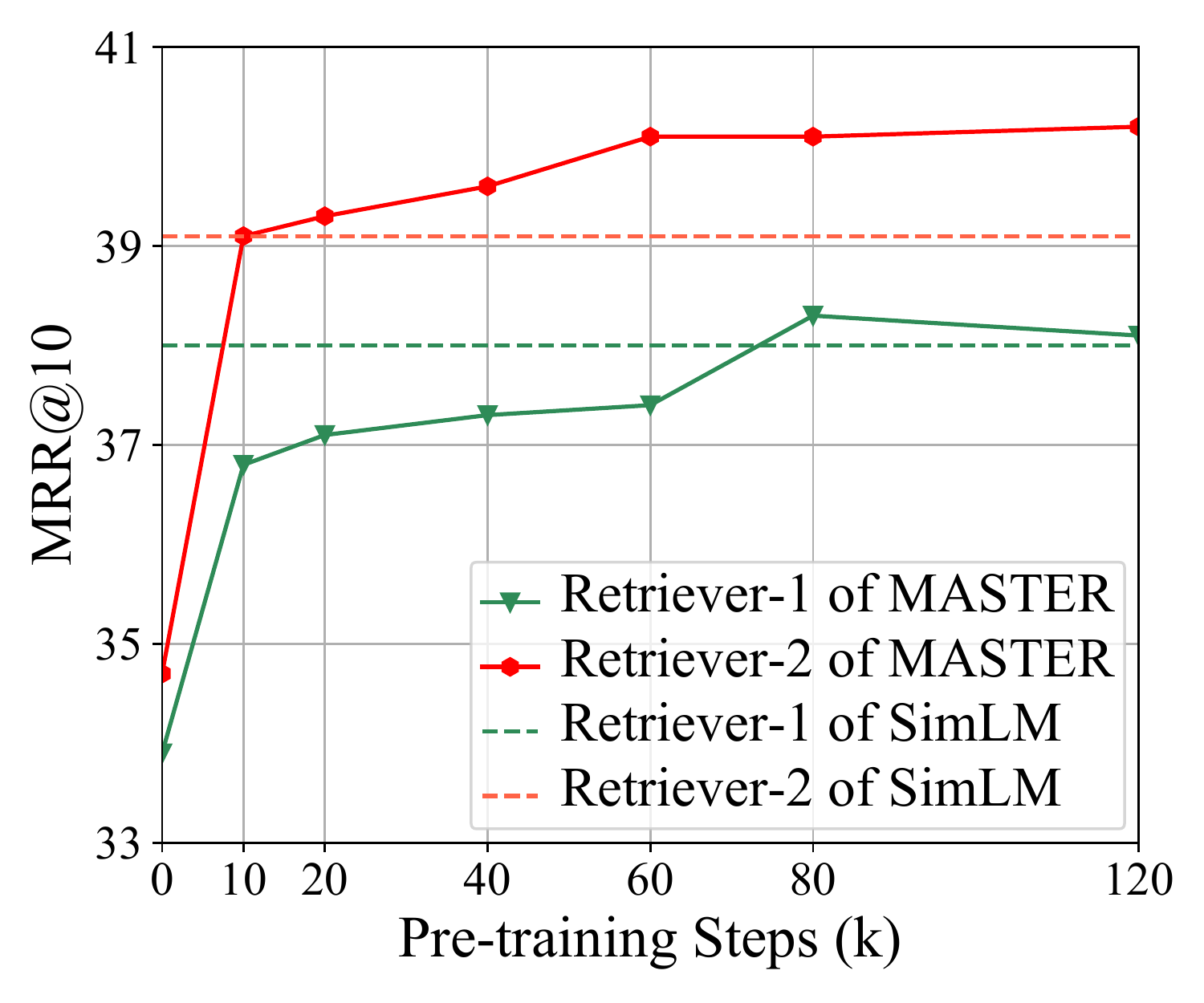}}
	\subfigure{\includegraphics[width=0.48\textwidth]{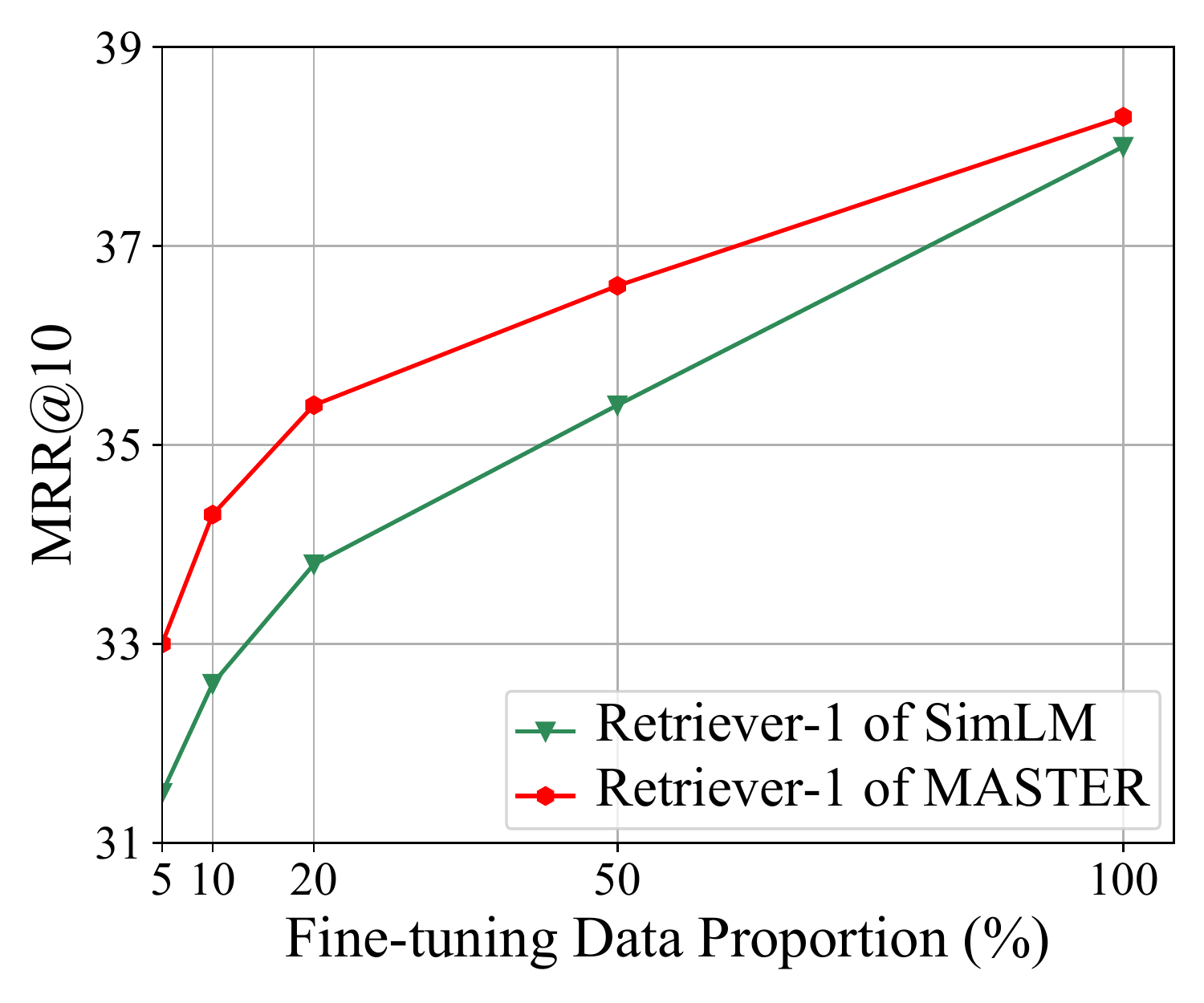}}
	\caption{
		Performance comparison w.r.t. different number of pre-training steps and data proportions on MS-MARCO.
	}\label{fig:training}
\end{figure}

\paratitle{Performance w.r.t. Different Pre-training Steps.}
As a pre-training approach, the number of pre-training steps will affect the performance on downstream tasks.
In each step, we optimize the model parameters using a batch of pre-training data by gradient descent algorithm.
However, too many pre-training steps are time-consuming and costly.
Here, we investigate the performance convergence speed of our approach during pre-training.
As shown in \cref{fig:training}(a), we can see that our model performs well with few pre-training steps, especially that the retriever$_2$ of our method achieves the 39.1 on MRR@10 metric (the same as SimLM) after 10k steps.
It shows that our approach is more effective to pre-train effective dense vectors, with no need for too many pre-training steps.

\ignore{
\begin{table}[t]
\small
\centering
\caption{Performance comparison (MRR@10) w.r.t. different training data size on the dev set of \textsc{MS-MARCO}.}
\begin{tabular}{l|ccccc}
\hline
 \textbf{Model} & \textbf{5\%} & \textbf{10\%} & \textbf{20\%} & \textbf{50\%} & \textbf{100\%} \\ 
 \hline
SimLM & 31.5 & 32.6 & 33.8 & 35.4 & 38.0 \\ 
MASTER & 33.0 & 34.3 & 35.4 & 36.6 & \textbf{38.3} \\
\hline
\end{tabular}
\label{tab:fewshot}
\end{table}}

\paratitle{Few-Shot Learning.}
In our approach, as we have pre-trained the backbone via a multi-task manner, the pre-learned dense vectors can be easily adapted into downstream tasks with less data.
To validate it, we reduce the training data size into 50\%, 20\%, 10\% and 5\%, and compare the performance of our approach with the pre-training method SimLM.
As shown in \cref{fig:training}(b), we can see that the performance substantially drops when less training data is used.
Additionally, our approach is consistently better than SimLM in all cases, especially in an extreme sparsity level (5\%).
It indicates that MASTER is better pre-trained to effectively adapt to downstream dense retrieval task.

\begin{table}[t]
\small
\centering
\caption{Experimental results on four NLU tasks from GLUE.}
\begin{tabular}{l|cccc}
\hline
\textbf{Model} & \textbf{CoLA} & \textbf{MRPC} & \textbf{STS-B} & \textbf{QQP} \\ \hline
BERT & 59.1 & 87.7 & 87.8 & 89.7  \\
Ours & 60.7 & 89.1 & 88.0 & 89.8  \\ \hline
\end{tabular}
\label{tab:NLU}
\end{table}

\paratitle{Natural Language Understanding Tasks.}
In our approach, as we integrate multiple pre-training tasks, our model can capture diverse knowledge from these tasks.
In this part, we evaluate if our pre-training methods can also benefit for natural language understanding (NLU) tasks.
We select the single-sentence and similarity tasks from the GLUE benchmark~\cite{DBLP:conf/iclr/WangSMHLB19} (\ie CoLA, MRPC, STS-B and QQP), which focus on predicting the acceptability, similarity and paraphrase of sentences from different domains (e.g., news and misc).
We fine-tune our pre-trained model on these tasks. and all the hyper-parameters are following the suggestions of the original BERT paper~\cite{devlin2019bert}.
As shown in Table~\ref{tab:NLU}, our approach improves the performance of BERT on these NLU tasks. It indicates that our multi-task pre-training can also enrich the useful knowledge about NLU tasks. 

\begin{table}[t]
\small
\centering
\caption{Performance comparison w.r.t. different masked rates in the encoder and decoder. We report MRR@10 of the Retriever$_1$ and Retriever$_2$ on MS-MARCO.}
\resizebox{\textwidth}{!}{\begin{tabular}{l|c|cc|cc}
\hline
\textbf{Model} & \textbf{30\% En-50\% De} & \textbf{15\% En-50\% De} & \textbf{50\% En-50\% De} & \textbf{30\% En-30\% De} & \textbf{30\% En-70\% De}  \\ \hline
Retriever$_{1}$ & \textbf{38.3} & 37.9 & 37.6 & 37.5 & 38.0  \\
Retriever$_{2}$ & \textbf{40.4} & 39.9 & 39.7 & 39.8 & 39.9  \\ \hline
\end{tabular}}
\label{tab:ptuning}
\end{table}

\paratitle{Hyper-parameter Tuning.}
The masked rates of the deep encoder and multiple decoders are two important hyper-parameters, as they control the information bottleneck in our approach.
Here, we set the masked rate in the encoder to be 15\%, 30\% and 50\%, and that in decoders to be 30\%, 50\% and 70\%.
\cref{tab:ptuning} shows the evaluation results.
First, our model is robust to these different settings.
Besides, when the masked rates of the encoder and decoders are set to 30\% and 50\% respectively, our model performs slightly better than others.
Therefore, we apply 30\% and 50\% as the masked rates of the encoder and decoders.

\section{Conclusion}
In this paper, we proposed MASTER, a multi-task pre-trained bottlenecked masked autoencoder for dense retrieval task.
In our approach, we adopted a bottlenecked multi-decoder architecture to integrate a variety of pre-training tasks, and devised three types of pre-training tasks about corrupted passages recovering, related passage recovering and PLMs outputs recovering.
The three types of tasks focused on compressing the information within the passages, modeling relations among passages, and learning the knowledge from external public generative PLMs, respectively, leading to more informative and effective dense vectors.
Experimental results have shown the superiority of our approach. 


\section*{Limitations}
A major limitation of our approach is the cost of pre-training.
Actually, it is not necessary for researchers or developers to complete the whole pre-training process, as they can directly utilize our publicly released checkpoints for initialization.
Besides, in this work, we evaluate our approach mainly on passage retrieval tasks, and do not consider the retrieval of very long documents.
Another possible issue derives from that we continually pre-train the parameters of BERT.
Since existing works have revealed that BERT might represent biases from the pre-training corpus, such an issue may also be inherited by our approach.

\section*{Acknowledgement}
Kun Zhou, Wayne Xin Zhao and Ji-Rong Wen were partially supported by National Natural Science Foundation of China under Grant No. 62222215, Beijing Natural Science Foundation under Grant No. 4222027, Beijing Outstanding Young Scientist Program under Grant No. BJJWZYJH012019100020098, and the Outstanding Innovative Talents Cultivation Funded Programs 2021 of Renmin University of China.
Wayne Xin Zhao is the corresponding author.
%
%
%
\bibliographystyle{splncs04}
\bibliography{custom}

\begin{thebibliography}{10}
\providecommand{\url}[1]{\texttt{#1}}
\providecommand{\urlprefix}{URL }
\providecommand{\doi}[1]{https://doi.org/#1}

\bibitem{DBLP:journals/corr/abs-2102-07662}
Craswell, N., Mitra, B., Yilmaz, E., Campos, D.: Overview of the {TREC} 2020
  deep learning track. CoRR  \textbf{abs/2102.07662} (2021),
  \url{https://arxiv.org/abs/2102.07662}

\bibitem{DBLP:journals/corr/abs-2003-07820}
Craswell, N., Mitra, B., Yilmaz, E., Campos, D., Voorhees, E.M.: Overview of
  the {TREC} 2019 deep learning track. CoRR  \textbf{abs/2003.07820} (2020),
  \url{https://arxiv.org/abs/2003.07820}

\bibitem{dai2019deeper}
Dai, Z., Callan, J.: Deeper text understanding for {IR} with contextual neural
  language modeling. In: Proceedings of {SIGIR} 2019. pp. 985--988 (2019),
  \url{https://doi.org/10.1145/3331184.3331303}

\bibitem{devlin2019bert}
Devlin, J., Chang, M.W., Lee, K., Toutanova, K.: {BERT}: Pre-training of deep
  bidirectional transformers for language understanding. In: Proceedings of
  {NAACL} 2019. pp. 4171--4186 (2019), \url{https://aclanthology.org/N19-1423}

\bibitem{Gao2021CondenserAP}
Gao, L., Callan, J.: Condenser: a pre-training architecture for dense
  retrieval. In: Proceedings of {EMNLP} 2021. pp. 981--993 (2021),
  \url{https://aclanthology.org/2021.emnlp-main.75}

\bibitem{gao2021your}
Gao, L., Callan, J.: Is your language model ready for dense representation
  fine-tuning? CoRR  \textbf{abs/2104.08253} (2021),
  \url{https://arxiv.org/abs/2104.08253}

\bibitem{DBLP:conf/acl/GaoC22}
Gao, L., Callan, J.: Unsupervised corpus aware language model pre-training for
  dense passage retrieval. In: Proceedings of {ACL} 2022. pp. 2843--2853
  (2022), \url{https://doi.org/10.18653/v1/2022.acl-long.203}

\bibitem{Gao2021COILRE}
Gao, L., Dai, Z., Callan, J.: {COIL}: Revisit exact lexical match in
  information retrieval with contextualized inverted list. In: Proceedings of
  {NAACL} 2021. pp. 3030--3042 (2021),
  \url{https://aclanthology.org/2021.naacl-main.241}

\bibitem{he2022masked}
He, K., Chen, X., Xie, S., Li, Y., Doll{\'a}r, P., Girshick, R.: Masked
  autoencoders are scalable vision learners. In: Proceedings of the IEEE/CVF
  Conference on Computer Vision and Pattern Recognition. pp. 16000--16009
  (2022)

\bibitem{DBLP:conf/sigir/HofstatterLYLH21}
Hofst{\"{a}}tter, S., Lin, S., Yang, J., Lin, J., Hanbury, A.: Efficiently
  teaching an effective dense retriever with balanced topic aware sampling. In:
  Proceedings of {SIGIR} 2021. pp. 113--122 (2021),
  \url{https://doi.org/10.1145/3404835.3462891}

\bibitem{DBLP:journals/tbd/JohnsonDJ21}
Johnson, J., Douze, M., J{\'{e}}gou, H.: Billion-scale similarity search with
  gpus. {IEEE} Transaction's on Big Data  \textbf{7}(3),  535--547 (2021),
  \url{https://doi.org/10.1109/TBDATA.2019.2921572}

\bibitem{DBLP:conf/emnlp/KarpukhinOMLWEC20}
Karpukhin, V., Oguz, B., Min, S., Lewis, P., Wu, L., Edunov, S., Chen, D., Yih,
  W.t.: Dense passage retrieval for open-domain question answering. In:
  Proceedings of {EMNLP} 2020. pp. 6769--6781 (2020),
  \url{https://aclanthology.org/2020.emnlp-main.550}

\bibitem{DBLP:conf/sigir/KhattabZ20}
Khattab, O., Zaharia, M.: Colbert: Efficient and effective passage search via
  contextualized late interaction over {BERT}. In: Proceedings of {SIGIR} 2020.
  pp. 39--48 (2020), \url{https://doi.org/10.1145/3397271.3401075}

\bibitem{DBLP:journals/tacl/KwiatkowskiPRCP19}
Kwiatkowski, T., Palomaki, J., Redfield, O., Collins, M., Parikh, A., Alberti,
  C., Epstein, D., Polosukhin, I., Devlin, J., Lee, K., Toutanova, K., Jones,
  L., Kelcey, M., Chang, M.W., Dai, A.M., Uszkoreit, J., Le, Q., Petrov, S.:
  Natural questions: A benchmark for question answering research. Transactions
  of the Association for Computational Linguistics  \textbf{7},  452--466
  (2019), \url{https://aclanthology.org/Q19-1026}

\bibitem{lee2019latent}
Lee, K., Chang, M.W., Toutanova, K.: Latent retrieval for weakly supervised
  open domain question answering. In: Proceedings of {ACL} 2019. pp. 6086--6096
  (2019), \url{https://aclanthology.org/P19-1612}

\bibitem{lin2023prod}
Lin, Z., Gong, Y., Liu, X., Zhang, H., Lin, C., Dong, A., Jiao, J., Lu, J.,
  Jiang, D., Majumder, R., et~al.: Prod: Progressive distillation for dense
  retrieval. In: Proceedings of the ACM Web Conference 2023. pp. 3299--3308
  (2023)

\bibitem{Liu2022RetroMAEPR}
Liu, Z., Shao, Y.: Retromae: Pre-training retrieval-oriented transformers via
  masked auto-encoder. CoRR  \textbf{abs/2205.12035} (2022),
  \url{https://doi.org/10.48550/arXiv.2205.12035}

\bibitem{Lu2021LessIM}
Lu, S., He, D., Xiong, C., Ke, G., Malik, W., Dou, Z., Bennett, P., Liu, T.Y.,
  Overwijk, A.: Less is more: Pretrain a strong {S}iamese encoder for dense
  text retrieval using a weak decoder. In: Proceedings of {EMNLP} 2021. pp.
  2780--2791 (2021), \url{https://aclanthology.org/2021.emnlp-main.220}

\bibitem{DBLP:journals/corr/abs-2205-09153}
Lu, Y., Liu, Y., Liu, J., Shi, Y., Huang, Z., Feng, S., Sun, Y., Tian, H., Wu,
  H., Wang, S., Yin, D., Wang, H.: Ernie-search: Bridging cross-encoder with
  dual-encoder via self on-the-fly distillation for dense passage retrieval.
  CoRR  \textbf{abs/2205.09153} (2022),
  \url{https://doi.org/10.48550/arXiv.2205.09153}

\bibitem{ma2023cot}
Ma, G., Wu, X., Wang, P., Hu, S.: Cot-mote: Exploring contextual masked
  auto-encoder pre-training with mixture-of-textual-experts for passage
  retrieval. arXiv preprint arXiv:2304.10195  (2023)

\bibitem{Ma2022PretrainAD}
Ma, X., Guo, J., Zhang, R., Fan, Y., Cheng, X.: Pre-train a discriminative text
  encoder for dense retrieval via contrastive span prediction. In: Proceedings
  of {SIGIR} 2022. pp. 848--858 (2022),
  \url{https://doi.org/10.1145/3477495.3531772}

\bibitem{DBLP:conf/nips/NguyenRSGTMD16}
Nguyen, T., Rosenberg, M., Song, X., Gao, J., Tiwary, S., Majumder, R., Deng,
  L.: {MS} {MARCO:} {A} human generated machine reading comprehension dataset.
  In: Proceedings of the Workshop on Cognitive Computation: Integrating neural
  and symbolic approaches 2016. vol.~1773 (2016),
  \url{http://ceur-ws.org/Vol-1773/CoCoNIPS\_2016\_paper9.pdf}

\bibitem{nogueira2019doc2query}
Nogueira, R., Lin, J.: From doc2query to doctttttquery  (2019),
  \url{https://cs.uwaterloo.ca/~jimmylin/publications/Nogueira_Lin_2019_docTTTTTquery.pdf}

\bibitem{DBLP:journals/corr/abs-1904-08375}
Nogueira, R.F., Yang, W., Lin, J., Cho, K.: Document expansion by query
  prediction. CoRR  \textbf{abs/1904.08375} (2019),
  \url{http://arxiv.org/abs/1904.08375}

\bibitem{Qu2021RocketQAAO}
Qu, Y., Ding, Y., Liu, J., Liu, K., Ren, R., Zhao, W.X., Dong, D., Wu, H.,
  Wang, H.: {R}ocket{QA}: An optimized training approach to dense passage
  retrieval for open-domain question answering. In: Proceedings of {NAACL}
  2021. pp. 5835--5847 (2021),
  \url{https://aclanthology.org/2021.naacl-main.466}

\bibitem{radford2019language}
Radford, A., Wu, J., Child, R., Luan, D., Amodei, D., Sutskever, I., et~al.:
  Language models are unsupervised multitask learners  (2019),
  \url{https://cdn.openai.com/better-language-models/language_models_are_unsupervised_multitask_learners.pdf}

\bibitem{raffel2020exploring}
Raffel, C., Shazeer, N., Roberts, A., Lee, K., Narang, S., Matena, M., Zhou,
  Y., Li, W., Liu, P.J.: Exploring the limits of transfer learning with a
  unified text-to-text transformer. Journal of Machine Learning Research
  \textbf{21},  140:1--140:67 (2020),
  \url{http://jmlr.org/papers/v21/20-074.html}

\bibitem{ramos2003using}
Ramos, J., et~al.: Using tf-idf to determine word relevance in document
  queries. In: Proceedings of the first instructional conference on machine
  learning. vol.~242, pp. 29--48 (2003)

\bibitem{DBLP:conf/acl/RenLQLZSWWW21}
Ren, R., Lv, S., Qu, Y., Liu, J., Zhao, W.X., She, Q., Wu, H., Wang, H., Wen,
  J.R.: {PAIR}: Leveraging passage-centric similarity relation for improving
  dense passage retrieval. In: Findings of the Association for Computational
  Linguistics: ACL-IJCNLP 2021. pp. 2173--2183 (2021),
  \url{https://aclanthology.org/2021.findings-acl.191}

\bibitem{DBLP:conf/emnlp/RenQLZSWWW21}
Ren, R., Qu, Y., Liu, J., Zhao, W.X., She, Q., Wu, H., Wang, H., Wen, J.:
  Rocketqav2: {A} joint training method for dense passage retrieval and passage
  re-ranking. In: Proceedings of {EMNLP} 2021. pp. 2825--2835 (2021),
  \url{https://doi.org/10.18653/v1/2021.emnlp-main.224}

\bibitem{DBLP:conf/naacl/SanthanamKSPZ22}
Santhanam, K., Khattab, O., Saad{-}Falcon, J., Potts, C., Zaharia, M.:
  Colbertv2: Effective and efficient retrieval via lightweight late
  interaction. In: Proceedings of the 2022 Conference of the North American
  Chapter of the Association for Computational Linguistics: Human Language
  Technologies, {NAACL} 2022, Seattle, WA, United States, July 10-15, 2022. pp.
  3715--3734 (2022), \url{https://doi.org/10.18653/v1/2022.naacl-main.272}

\bibitem{sun2022lead}
Sun, H., Liu, X., Gong, Y., Dong, A., Jiao, J., Lu, J., Zhang, Y., Jiang, D.,
  Yang, L., Majumder, R., et~al.: Lead: Liberal feature-based distillation for
  dense retrieval. arXiv preprint arXiv:2212.05225  (2022)

\bibitem{DBLP:journals/corr/abs-2104-08663}
Thakur, N., Reimers, N., R{\"{u}}ckl{\'{e}}, A., Srivastava, A., Gurevych, I.:
  {BEIR:} {A} heterogenous benchmark for zero-shot evaluation of information
  retrieval models. CoRR  \textbf{abs/2104.08663} (2021),
  \url{https://arxiv.org/abs/2104.08663}

\bibitem{DBLP:conf/iclr/WangSMHLB19}
Wang, A., Singh, A., Michael, J., Hill, F., Levy, O., Bowman, S.R.: {GLUE:} {A}
  multi-task benchmark and analysis platform for natural language
  understanding. In: Proceedings of {ICLR} 2019 (2019),
  \url{https://openreview.net/forum?id=rJ4km2R5t7}

\bibitem{DBLP:journals/corr/abs-2207-02578}
Wang, L., Yang, N., Huang, X., Jiao, B., Yang, L., Jiang, D., Majumder, R.,
  Wei, F.: Simlm: Pre-training with representation bottleneck for dense passage
  retrieval. CoRR  \textbf{abs/2207.02578} (2022),
  \url{https://doi.org/10.48550/arXiv.2207.02578}

\bibitem{DBLP:journals/corr/abs-2208-07670}
Wu, X., Ma, G., Lin, M., Lin, Z., Wang, Z., Hu, S.: Contextual mask
  auto-encoder for dense passage retrieval. CoRR  \textbf{abs/2208.07670}
  (2022), \url{https://doi.org/10.48550/arXiv.2208.07670}

\bibitem{wu2023cot}
Wu, X., Ma, G., Wang, P., Lin, M., Lin, Z., Zhang, F., Hu, S.: Cot-mae v2:
  Contextual masked auto-encoder with multi-view modeling for passage
  retrieval. arXiv preprint arXiv:2304.03158  (2023)

\bibitem{xiao2022retromae}
Xiao, S., Liu, Z.: Retromae v2: Duplex masked auto-encoder for pre-training
  retrieval-oriented language models. arXiv preprint arXiv:2211.08769  (2022)

\bibitem{DBLP:conf/iclr/XiongXLTLBAO21}
Xiong, L., Xiong, C., Li, Y., Tang, K., Liu, J., Bennett, P.N., Ahmed, J.,
  Overwijk, A.: Approximate nearest neighbor negative contrastive learning for
  dense text retrieval. In: 9th International Conference on Learning
  Representations, {ICLR} 2021, Virtual Event, Austria, May 3-7, 2021 (2021),
  \url{https://openreview.net/forum?id=zeFrfgyZln}

\bibitem{yang2017anserini}
Yang, P., Fang, H., Lin, J.: Anserini: Enabling the use of lucene for
  information retrieval research. In: Proceedings of the 40th International
  {ACM} {SIGIR} Conference on Research and Development in Information
  Retrieval, Shinjuku, Tokyo, Japan, August 7-11, 2017. pp. 1253--1256 (2017),
  \url{https://doi.org/10.1145/3077136.3080721}

\bibitem{DBLP:conf/sigir/ZhanM0G0M21}
Zhan, J., Mao, J., Liu, Y., Guo, J., Zhang, M., Ma, S.: Optimizing dense
  retrieval model training with hard negatives. In: {SIGIR} '21: The 44th
  International {ACM} {SIGIR} Conference on Research and Development in
  Information Retrieval, Virtual Event, Canada, July 11-15, 2021. pp.
  1503--1512 (2021), \url{https://doi.org/10.1145/3404835.3462880}

\bibitem{DBLP:conf/iclr/ZhangGS0DC22}
Zhang, H., Gong, Y., Shen, Y., Lv, J., Duan, N., Chen, W.: Adversarial
  retriever-ranker for dense text retrieval. In: The Tenth International
  Conference on Learning Representations, {ICLR} 2022, Virtual Event, April
  25-29, 2022 (2022), \url{https://openreview.net/forum?id=MR7XubKUFB}

\bibitem{zhao2022dense}
Zhao, W.X., Liu, J., Ren, R., Wen, J.R.: Dense text retrieval based on
  pretrained language models: A survey. arXiv preprint arXiv:2211.14876  (2022)

\bibitem{zhao2023survey}
Zhao, W.X., Zhou, K., Li, J., Tang, T., Wang, X., Hou, Y., Min, Y., Zhang, B.,
  Zhang, J., Dong, Z., et~al.: A survey of large language models. arXiv
  preprint arXiv:2303.18223  (2023)

\bibitem{zhou2022simans}
Zhou, K., Gong, Y., Liu, X., Zhao, W.X., Shen, Y., Dong, A., Lu, J., Majumder,
  R., Wen, J.R., Duan, N., et~al.: Simans: Simple ambiguous negatives sampling
  for dense text retrieval. In: Proceedings of the 2022 Conference on Empirical
  Methods in Natural Language Processing (EMNLP) (2022)

\bibitem{zhou2022debiased}
Zhou, K., Zhang, B., Zhao, W.X., Wen, J.R.: Debiased contrastive learning of
  unsupervised sentence representations. In: Proceedings of the 60th Annual
  Meeting of the Association for Computational Linguistics (Volume 1: Long
  Papers). pp. 6120--6130 (2022)

\bibitem{zhou2023dynamicretriever}
Zhou, Y.J., Yao, J., Dou, Z.C., Wu, L., Wen, J.R.: Dynamicretriever: A
  pre-trained model-based ir system without an explicit index. Machine
  Intelligence Research pp. 1--13 (2023)

\end{thebibliography}

\newpage
\appendix

\begin{table*}[t]
\centering
\small
\begin{tabular}{p{3cm}|p{9cm}}
\toprule
\hline
Query & perveance definition  \\ \midrule
Top-1 by BERT & Definition of Pervade. transitive verb To pass or flow through, as an aperture, pore, or interstice; to permeate. \\ \midrule
Top-1 by PT w CPR & Perveance. Perveance is a notion used in the description of charged particle beams. The value of perveance indicates how significant the space charge effect is on the beam motion. \\ \midrule  
\bottomrule
Query & when typing which formula is used to measure accuracy  \\ \midrule
Top-1 by PT w CPR & Formula for Measure Typing Speed in Mastering Typing. Include gross WPM and net WPM that used to calculate accuracy in Mastering Typing Formula for Measure Typing Speed in Mastering Typing. \\ \midrule
Top-1 by PT w CPR+RPR & Formula to measure typing speed in WPM. Accuracy Formula: Accuracy is a percentage ratio of Gross and Net Word Speed: Accuracy = Net-WPM / Gross-WPM * 100. Calculation of Errors. Errors are calculated by following two criteria. Errors that are made and corrected; Errors that are made and not corrected. \\ \midrule  
\bottomrule
Query & what does dress it down mean \\ \midrule
Top-1 by PT w CPR+RPR & When people say dress down/up what does it mean?. Asker's rating. 1  Dress Down Meaning. 2  I can nest answer this with an example. 3  to dress it up means to go more formal than school/work clothes. 4  Dressing up is when you are doing something fancier then an everyday thing. 5  For the best answers, search on this site. \\ \midrule
Top-1 by PT w CPR+RPR+POR & When people say dress down/up what does it mean?. Best Answer: Dress up means to dress more fancier, like you are going out to a fancy restaurant and you want to dress up for it. To dress down, means if you are wearing something fancy and you are going somewhere that doesn't need you to be so well dressed, you dress down into something more casual. \\ \midrule  
\bottomrule
\end{tabular}	
    \caption{Examples of retrieved results on MS-MARCO by our approach.}	
    \label{tab-case}
\end{table*}

\section{Case Study}
To further analyze the effectiveness of our proposed three types of pre-training tasks, we select representative examples from the MS-MARCO dataset, to show the Win/Lose case study of the retrieved top-1 document using different pre-training objectives.
As shown in Table~\ref{tab-case}, from the first example, we can see that given the query about the definition of perveance, the original BERT has retrieved the definition of ``pervade'' in the 1st place by mistake. As discussed in Condenser~\cite{Gao2021CondenserAP}, the internal attention structure of BERT is not ready-to-use for dense encoders, which fails to aggregate the information of the keyword ``perveance'' into the dense representation. As a comparison, after pre-training with CPR that focuses on modeling the semantic information within the passage, our model successfully retrieves the relevant document.

From the second example, we can see that the retrieved documents from the two variations are indeed relevant to the given query, and many words are co-occurring in the query and documents. Whereas, the retrieved one from PT w CPR+RPR is more proper to answer the given query, since it focuses on the keyword "measure accuracy" and provides a detailed description of the "Accuracy Formula" to illustrate how it can measure the accuracy. It indicates that with the help of the RPR task, our model can better focus on useful information from the text that can capture important relationships between queries and passages.

From the third example, we can see that the two retrieved documents are very relevant to the given query and both focus on the keyword "dress down". However, the retrieved one from PT w CPR+RPR does not clearly answer the given query but just lists several possible answers. As a comparison, the retrieved one from PT w CPR+RPR+POR provides a concrete answer about the means of "dress down". Since POR loss aims to learn the correlations between the passage and the PLM-generated related text, it can meet more examples from other publicly released PLMs, further improving its capacity to model complex query-document relationships.

\end{document}